\documentclass[11pt]{article}
\usepackage{color}
\usepackage{algorithm, algpseudocode}
\usepackage{mathrsfs}
\usepackage{dsfont}
\usepackage{lmodern}
\usepackage{array}
\usepackage{bbm}
\usepackage{amsmath}
\usepackage{amssymb}
\usepackage[mathscr]{eucal}
\usepackage{graphicx}
\usepackage{mathrsfs}
\usepackage{psfrag}
\usepackage{color}
\usepackage{here}
\usepackage[margin=1in]{geometry}
\usepackage{wasysym}
\usepackage{enumitem}
\usepackage{xcolor}
\usepackage{caption}
\usepackage{subcaption}
\usepackage{amsthm}
\usepackage{lipsum}
\usepackage[round]{natbib}

\definecolor{darkblue}{RGB}{0,0,140}
\usepackage[colorlinks=true,
            allcolors=darkblue]{hyperref}

\newtheorem{lemma} {Lemma}

\newtheorem{example} {Example}

\newcommand{\Exp}{\mathds{E}}

\newcommand{\Prob}{\mathds{P}}

\newcommand{\Nat}{\mathbb{N}}
\newcommand{\Ind}{\mathds{1}}

\DeclareMathOperator*{\argmax}{arg\,max}

\newcommand{\Fc}{\mathcal{F}}

\newcommand{\Sc}{\mathcal{S}}
\newcommand{\Ac}{\mathcal{A}}
\newcommand{\Mc}{\mathcal{M}}

\newcommand{\Hc}{\mathcal{H}}

\title{Posterior Sampling for Reinforcement Learning Without Episodes}

\author{
Ian Osband \\
Stanford University, Google DeepMind\\
\texttt{iosband@stanford.edu}
\and
Benjamin Van Roy \\
Stanford University\\
\texttt{bvr@stanford.edu}
}
\begin{document}
\maketitle

%%%%%%%%%%%%%%%%%%%%%%%%%%%%%%%%%%%%%%%%%%%%%%%%%%%%%%%%%%%%%%%%%%%%%%%%%%%
%%%%%%%%%%%%%%%%%%%%%%%%%%%%%%%%%%%%%%%%%%%%%%%%%%%%%%%%%%%%%%%%%%%%%%%%%%%
\section{Introduction}

This is a brief technical note to clarify some of the issues with applying the application of the algorithm \textit{posterior sampling for reinforcement learning} (PSRL) in environments without fixed episodes.
In particular, this paper aims to:
\begin{itemize}
     \item Review some of results which have been proven for finite horizon MDPs \cite{Osband2013,osband2014near,osband2014model,osband2016posterior} and also for MDPs with finite ergodic structure \cite{gopalan2014thompson}.

     \item Review similar results for optimistic algorithms in infinite horizon problems \cite{Jaksch2010,Bartlett2009,abbasi2011regret}, with particular attention to the dynamic episode growth.

     \item Highlight the delicate technical issue which has led to a fault in the proof of the lazy-PSRL algorithm \cite{abbasi2015bayesian}.
     We present an explicit counterexample to this style of argument.
     Therefore, we suggest that the Theorem 2 in \cite{abbasi2015bayesian} be instead considered a conjecture, as it has no rigorous proof.

     \item Present pragmatic approaches to apply PSRL in infinite horizon problems. We conjecture that, under some additional assumptions, it will be possible to obtain bounds $O( \sqrt{T} )$ even without episodic reset.

\end{itemize}

%%%%%%%%%%%%%%%%%%%%%%%%%%%%%%%%%%%%%%%%%%%%%%%%%%%%%%%%%%%%%%%%%%%%%%%%%%%
%%%%%%%%%%%%%%%%%%%%%%%%%%%%%%%%%%%%%%%%%%%%%%%%%%%%%%%%%%%%%%%%%%%%%%%%%%%
\section{Problem formulation}
\label{sec: prob form}

We consider the problem of learning to optimize an unknown MDP $M^* = (\Sc, \Ac, R^*, P^*)$.
$\Sc = \{1,..,S\}$ is the state space, $\Ac=\{1,..,A\}$ is the action space.
In each timestep $t=1,2,..$ the agent observes a state $s_t \in \Sc$, selects an action $a_t \in \Ac$, receives a reward $r_t \sim R^*(s_t,a_t) \in [0,1]$ and transitions to a new state $s_{t+1} \sim P^*(s_t, a_t)$.
We define all random variables with respect to a probability space $(\Omega, \Fc, \Prob)$.

A policy $\mu$ is a mapping from state $s \in \Sc$ to action $a \in \Ac$.
For MDP $M$ and any policy $\mu$ we define the long run average reward starting from state $s$:
\begin{equation}
    \lambda^M_\mu(s) := \lim_{T \rightarrow \infty}
    \Exp_{M, \mu} \left[ \frac{1}{T} \sum_{t=1}^T \overline{r}(s_t,a_t) \ \mid \ s_1 = s \right],
\end{equation}
where $\overline{r}^*(s,a) := \Exp[r | r \sim R^*(s,a)]$.
The subscripts $M, \mu$ indicate the MDP evolves under $M$ with policy $\mu$.
A policy $\mu^M$ is optimal for the MDP $M$ if $\mu^M \in \arg\max_\mu \lambda^M_\mu(s)$ for all $s \in \Sc$.
For the unknown MDP $M^*$ we will often abbreviate sub/superscripts to simply $*$, for example $\lambda^*_*$ for $\lambda^{M^*}_{\mu^{M^*}}$.

Let $\Hc_t=(s_1, a_1, r_1,..,s_{t-1}, a_{t-1}, r_{t-1})$ denote the history of observations made \textit{prior} to time $t$.
A reinforcement learning algorithm is a deterministic sequence $\{ \pi_t | t=1,2,.. \}$ of functions each mapping $\Hc_t$ to a probability distribution $\pi_t(\Hc_t)$ over policies, from which the agent sample policy $\mu_t$ at timestep $t$.
We define the regret of a reinforcement learning algorithm $\pi$ up to time $T$
\begin{equation}
    {\rm Regret}(T, \pi, M^*)(s) := \sum_{t=1}^T \left\{ \lambda^*_*(s) - r_t \right\} \bigg| s_1 = s.
\end{equation}

The regret of a learning algorithm shows how worse the policy performs that optimal in terms of cumulative rewards.
Any algorithm with $o(T)$ regret will \textit{eventually} learn the optimal policy.
Note that the regret is random since it depends on the unknown MDP $M^*$, the random sampling of policies and, through the history $\Hc_t$ on the previous transitions and rewards.
We will assess and compare algorithm performance in terms of the regret.

\subsection{Finite horizon MDPs}

We now spend a little time to relate the formulation above to so-called finite horizon MDPs \cite{Osband2013,dann2015sample}.
In this setting, an agent will interact repeatedly with a environment over $H \in \Nat$ timesteps which we call an \textit{episode}.
A finite horizon MDP $M^* = (\Sc, \Ac, R^*, P^*, H, \rho)$ is defined as above, but every $H \in \Nat$ timesteps the state will reset according to some initial distribution $\rho$.
We call $H \in \Nat$ the horizon of the MDP.

In a finite horizon MDP a typical policy may depend on both the state $s \in \Sc$ and the timestep $h$ within the episode.
To be explicit, we define a policy $\mu$ is a mapping from state $s \in \Sc$ and period $h=1,..,H$ to action $a \in \Ac$.
For each MDP $M = (\Sc, \Ac, R^M \hspace{-1mm}, P^M\hspace{-1mm}, H, \rho)$ and policy $\mu$ we define the state-action value function for each period $h$:
\vspace{-1mm}
\begin{equation}
\label{eq: q value tabular}
  Q^{M}_{\mu, h}(s, a) := \Exp_{M,\mu}\left[ \sum_{j=h}^{H} \overline{r}^M(s_j,a_j) \Big| s_h = s, a_h=a \right],
\end{equation}
and $V^{M}_{\mu, h}(s) := Q^M_{\mu, h}(s, \mu(s,h))$.
Once again, we say a policy $\mu^M$ is optimal for the MDP $M$ if $\mu^M \in \argmax_{\mu} V^{M}_{\mu, h}(s)$ for all $s \in \Sc$ and $h=1,\ldots,H$.

At first glance this might seem at odds with the formulation in Section \ref{sec: prob form}.
However, finite horizon MDPs can be thought of as a special case of Section \ref{sec: prob form} in the expanded state space $\tilde{\Sc} := \Sc \times \{1,..,H\}$.
In this case it is typical to assume that the agent \textit{knows} about the evolution of time $h$ deterministically a priori.
To highlight this time evolution within episodes, with some abuse of notation, we let $s_{kh} = s_t$ for $t=(k-1)H+h$, so that
$s_{kh}$ is the state in period $h$ of episode $k$.  We define $\Hc_{kh}$ analogously.

%%%%%%%%%%%%%%%%%%%%%%%%%%%%%%%%%%%%%%%%%%%%%%%%%%%%%%%%%%%%%%%%%%%%%%%%%%%
%%%%%%%%%%%%%%%%%%%%%%%%%%%%%%%%%%%%%%%%%%%%%%%%%%%%%%%%%%%%%%%%%%%%%%%%%%%
\section{Posterior sampling for reinforcement learning}
\label{sec: psrl}

The algorithm posterior sampling for reinforcement learning (PSRL) was first proposed as \textit{Bayesian dynamic programming} by \cite{Strens00}.
Later work has altered the terminology to highlight the distinction between PSRL and the Bayes-optimal solution which can be found by dynamic programming in the Bayesian belief state \cite{Osband2013}.
PSRL begins with a prior distribution $\phi$ over MDPs.
At the start of each $k$th episode, PSRL samples an MDP $M_k$ from this posterior belief and follows the policy which is optimal for that \textit{sample} over episode $k$.
We specify this process in Algorithm \ref{alg: PSRL}.

\begin{algorithm}[H]
\caption{\protect PSRL}
\label{alg: PSRL}
{\small
\begin{algorithmic}[1]
\State{\textbf{Input:} prior distribution $\phi$, episode length $H$}
    \For{episode $k=1,2,..$}
        \State{sample MDP $M_k \ \sim \ \phi(\cdot | \Hc_{k1})$}
        \State{compute $\mu_k \ \in \  \argmax\limits_{\mu} V^{M_k}_{\mu, 1}$}
    \For{time $h = 1,2,..,H$ }
        \State{take action $a_{kh} = \mu_k(s_{kh}, h)$}
        \State{observe $r_{kh}$ and $s_{kh+1}$}
        \State{update $\Hc_{kh} = \Hc_{kh} \ \cup \ (a_{kh}, r_{kh}, s_{kh+1})$}
    \EndFor
    \EndFor
\end{algorithmic}
}
\end{algorithm}

PSRL is a general algorithmic approach and can be used with any possible prior $\phi$ over MDPs.
Prior knowledge to the structure and generalization of the MDP can be encoded in $\phi$ in an arbitrary way.
For MDPs with finite states and actions and little prior knowledge, it may be natural to use an uninformative conjugate prior for the rewards and transitions in each state and action independently.
A simple implementation of this type of PSRL is available at \url{https://github.com/iosband/TabulaRL}.

We note that, in the statement of Algorithm \ref{alg: PSRL} the episode length $H$ is given to the agent a priori.
In its standard implementation, PSRL proceeds in fixed episodes of length $H$.
For finite horizon MDPs $H$ is typically given to be the horizon of the MDP.
For MDPs without episodic reset, PSRL imposes artificial episodes of fixed policies even though the underlying dynamics of the MDP are time-homogeneous.

%%%%%%%%%%%%%%%%%%%%%%%%%%%%%%%%%%%%%%%%%%%%%%%%%%%%%%%%%%%%%%%%%%%%%%%%%%%
%%%%%%%%%%%%%%%%%%%%%%%%%%%%%%%%%%%%%%%%%%%%%%%%%%%%%%%%%%%%%%%%%%%%%%%%%%%
\subsection{Regret bounds in episodic environments}
\label{sec: episodic}

Recently, several papers have established performance guarantees for PSRL in finite horizon MDPs \cite{Osband2013,osband2014near,osband2014model,osband2016posterior}.
For all of these papers, the key observation comes from earlier work \cite{Russo2013} and the posterior sampling lemma.

\begin{lemma}[Posterior sampling]
\label{lem: posterior_sampling}
\hspace{0.000001mm} \newline
If $\phi$ is the distribution of $M^*$ then for any $\sigma(\Hc_{k1})$-measurable function $g$,
\begin{equation}
\label{eq: ps}
    \Exp[g(M^*) | \Hc_{k1} ] = \Exp[g(M_k) | \Hc_{k1} ].
\end{equation}
Note that \eqref{eq: ps} also shows that $\Exp[g(M^*)] = \Exp[g(M_k)]$ through the tower property.
\end{lemma}

Most papers which bound the Bayesian regret for PSRL begin with the standard analysis of optimistic algorithms to add and subtract the \textit{imagined} optimal reward $V^{M_k}_{\mu_k, 1}$.
\begin{eqnarray}
    \Delta_k &:=& V^{M^*}_{\mu^*,1} - V^{M^*}_{\mu_k,1} \nonumber \\
    &=& \underbrace{V^{M^*}_{\mu^*,1} - V^{M_k}_{\mu_k,1}}_{\Delta^{\rm opt}_k}
     + \underbrace{V^{M_k}_{\mu_k,1}- V^{M^*}_{\mu_k,1}}_{\Delta^{\rm conc}_k}
\end{eqnarray}
Where $\Delta^{\rm opt}_k$ is the regret from optimism and $\Delta^{\rm conc}_k$ is the regret from concentration.
We then Lemma \ref{lem: posterior_sampling} to assert that, conditional on any past data \cite{Russo2013,Osband2013}
\begin{equation}
     \Exp[\Delta^{\rm opt}_k | \Hc_{k1}] = \Exp[ V^*_{*1} - V^k_{k1} | \Hc_{k1}] = 0.
\end{equation}
Crucially, the remaining term $\Delta^{\rm conc}_k$ only depends on the policy $\mu_k$ which PSRL \textit{actually} follows, and not the unknown optimal policy $\mu^*$ which is unobserved.
Standard proofs conclude with some concentration results which bound $\Delta^{\rm conc}_k$ \cite{Osband2013,osband2014near,osband2014model,osband2016posterior}.

A similar argument is used in the frequentist analysis of the PSRL variant studied in \cite{gopalan2014thompson}.
Although these authors do not explicitly consider finite horizon MDPs, they assume that the environment is ergodic and there exists some positive recurrent state $s_0$ under \textit{any} policy.
This assumption and the resulting analysis share many similarities with the finite horizon setting.

%%%%%%%%%%%%%%%%%%%%%%%%%%%%%%%%%%%%%%%%%%%%%%%%%%%%%%%%%%%%%%%%%%%%%%%%%%%
%%%%%%%%%%%%%%%%%%%%%%%%%%%%%%%%%%%%%%%%%%%%%%%%%%%%%%%%%%%%%%%%%%%%%%%%%%%
\section{Infinite horizon problems}
\label{sec: infinite}

Algorithms that bound the regret over MDPs without finite horizon must impose some connectedness constraint in order to guarantee regret $o(T)$.
To see why this is the case, consider the problem in Example \ref{ex: infinite regret}.

\begin{example}[Heaven and hell]
\label{ex: infinite regret}
\hspace{0.000001mm} \newline
We consider a simple MDP with three states $\Sc = \{s_0, s_1, s_2\}$ and two actions $\Ac = \{a_1, a_2\}$.
The agent begins in $s_0$, if they choose action $a_1$ they transition to $s_1$ and if they choose $a_2$ they transiton to $s_2$.
The states $s_1$ and $s_2$ are absorbing, so that for all subsequent steps they remain here, regardless of action.
One of these states is ``heaven'' in that the agent will receive the maximum reward of one for all time, the other one is ``hell'' and it will never get a reward again.
The problem is that the agent does not know which one is which.
Even if the agent knows the entire structure of the problem, if the optimal action $a^*$ is equally likely to be $1$ or $2$ there is no algorithm that can provide an expected regret less than $\frac{T}{2}$.
\end{example}

Clearly, problems of the style of Example \ref{ex: infinite regret} are not amenable to meaningful regret analysis.
In order to design an algorithm with sublinear regret we must restrict our attention to MDPs with some connected structure.
The precise requirements used in each paper are slightly different, but may include \cite{Bartlett2009},
{\small
\begin{itemize}
     \item \textbf{Ergodic}, it is possible to reach any state from any other state under any policy.

     \item \textbf{Unichain}, every policy induces a single recurrent class plus a set of transient states.

     \item \textbf{Communicating}, for every $s_1, s_2 \in \Sc$ there is \textit{some} policy that takes an agent from $s_1$ to $s_2$.

     \item \textbf{Weakly communicating}, The state space $\Sc$ decomposes into two sets: in the first, each state is reachable from every other state from every other state under some policy; in the second, all states are transient under all policies.
\end{itemize}}

In each of these settings some notion of MDP connectedness emerges which is somewhat comparable to the horizon $H$ in finite horizon MDPs.
However, and unlike the analysis of finite horizon MDPs, this connectedness parameters is typically not known to the agent a priori.
Many algorithms, including UCRL2 \cite{Jaksch2010} and REGAL \cite{Bartlett2009}, are able to learn with provable regret bounds without prior knowledge of the unknown MDP connectedness.
In Section \ref{sec: existing} we provide a sketch of how they are able to accomplish this through ``the doubling trick''.

%%%%%%%%%%%%%%%%%%%%%%%%%%%%%%%%%%%%%%%%%%%%%%%%%%%%%%%%%%%%%%%%%%%%%%%%%%%
%%%%%%%%%%%%%%%%%%%%%%%%%%%%%%%%%%%%%%%%%%%%%%%%%%%%%%%%%%%%%%%%%%%%%%%%%%%
\subsection{Existing optimistic analyses}
\label{sec: existing}

Many algorithms for efficient reinforcement learning in infinite horizon MDPs are driven by the principle of ``optimism in the face of uncertainty'' (OFU) \cite{munos2014bandits}.
At a high level, many of these algorithms fall into the general structure of Algorithm \ref{alg: OFU}.
These algorithms share a lot of similarities with PSRL, but instead of sampling a single MDP from the posterior and following the policy which is optimal for that sample, they build up a confidence set of plausible MDPs and then follow the policy which is most optimistic within all plausible MDPs.

\begin{algorithm}[H]
\caption{\protect OFU RL}
\label{alg: OFU}
{\small
\begin{algorithmic}[1]
\State{\textbf{Input:} confidence set constructor $\Phi$, episode signal $E$}
    \For{episode $k=1,2,..$}
        \State{construct confidence set $\Mc_k \ = \ \Phi(\Hc_{k1})$}
        \State{compute $\mu_k \ \in \ \argmax\limits_{\mu}\ \max\limits_{M \in \Mc_k} \lambda^M_{\mu}$}
    \For{while $E(\Hc_{kh}) = {\rm FALSE}$ }
        \State{take action $a_{kh} = \mu_k(s_{kh})$}
        \State{observe $r_{kh}$ and $s_{kh+1}$}
        \State{update $\Hc_{kh} = \Hc_{kh} \ \cup \ (a_{kh}, r_{kh}, s_{kh+1})$}
    \EndFor
    \EndFor
\end{algorithmic}
}
\end{algorithm}

Another difference from finite horizon MDPs is that the episode length of fixed policies is not necessarily fixed, but may depend on the data which is gathered by the algorithm through the episode signal $E(\Hc_{kh})$.
A common scheme, used in UCRL2 \cite{Jaksch2010} and REGAL \cite{Bartlett2009}, is to only start a new episode when the total number of visits to any state and action has doubled.
However, similar schemes which successively grow the length of the episodes under consideration show up in several settings \cite{abbasi2011regret}.
This kind of idea, which successively doubles the length of the policies under evaluation, is so pervasive in the RL literature that it is often simply referred to as ``the doubling trick''.

Algorithms for infinite horizon problems typically do not use episodes of fixed length $H$.
The reasons for this are as follows:
\begin{itemize}[noitemsep, nolistsep]
    \item If $H$ is smaller than the corresponding mixing time of the optimal policy the agent may not be able to \textit{ever} learn the optimal policy.

    \item If $H$ is much larger than the timeframe of the optimal policy then the agent may be overly wasteful in not updating its policies as data arrives.

    \item Every $H$ steps the agent may incur some nonzero opportunity cost from switching policies. If this occurs every $H$ steps then the algorithm will never guarantee sublinear regret.
\end{itemize}
The big picture idea of ``the doubling trick'' is to successively grow the length of the episodes under consideration (by doubling) to begin with short policies but increase their length as time goes on.
Importantly, because the growth in the length of policies is exponential, this will only contribute logarithmic switching costs as $T$ grows.

The analysis for Algorithm \ref{alg: OFU} is not particularly complicated by the introduction of data-dependent episode signalling $E$ since, by construction, the confidence set constructors generate high probability confidence sets $\Phi(\Hc_{k1})$ such that $\lambda^{M_k}_{\mu_k} \ge \lambda^*_*$for \textit{any} possible data $\Hc_{k1}$.
At a high level, and ignoring the mixing costs of the MDP at changes of policy (which are not very important when they only occur logarithmically in $T$), we can approximate the regret in any episode $k$ by the difference in long run optimal rewards:
$$ \Delta_k := L^*_k \left( \lambda^*_* - \lambda^*_k \right) \text{ where } L^*_k \text{ is the length of episode } k.$$

We continue the proof for Algorithm \ref{alg: OFU} in the standard way, by adding and subtracting the imagined optimal reward.
The optimistic principle means that, for any episode $k$ in which the true MDP $M^*$ lies within the confidence set $\Mc_k$ then $\lambda^k_k = \max\limits_{\mu, M \in \Mc_k} \lambda^M_\mu \ge \lambda^*_*$.
Since the confidence set $\Mc_k$ is designed to contain the true MDP with high probability then we can say for \textit{any} random episode length $L^*_K \ge 0$,
\begin{eqnarray}
    \Delta_k &:=& L^*_k \left(\lambda^*_* - \lambda^k_k \right) \nonumber \\
    &=&  \underbrace{L^*_k(\lambda^*_* - \lambda^k_k)}_{\Delta^{\rm opt}_k}
     + \underbrace{L^*_k(\lambda^k_k - \lambda^*_k)}_{\Delta^{\rm conc}_k} \nonumber \\
    &\le& \underbrace{L^*_k(\lambda^k_k - \lambda^*_k)}_{\Delta^{\rm conc}_k} \text{ with high probability.}
\end{eqnarray}
The proofs for Algorithm \ref{alg: OFU} conclude with standard concentration inequalities for $\Delta^{\rm conc}_k$ \cite{Jaksch2010,Bartlett2009,abbasi2011regret}.
At first glance, it looks like the same kind of analysis should immediately apply to a modified variant of PSRL, which resamples episodes according to $E(\cdot)$.
In fact, as we will show in the next section, this is not generally the case.

%%%%%%%%%%%%%%%%%%%%%%%%%%%%%%%%%%%%%%%%%%%%%%%%%%%%%%%%%%%%%%%%%%%%%%%%%%%
%%%%%%%%%%%%%%%%%%%%%%%%%%%%%%%%%%%%%%%%%%%%%%%%%%%%%%%%%%%%%%%%%%%%%%%%%%%
\subsection{The problem with PSRL}
\label{sec: lazy_psrl}

Recent analysis has attempted to extend the analysis of PSRL to the infinite horizon setting.
The proposed algorithm Lazy-PSRL \cite{abbasi2015bayesian}, modifies PSRL to use a dynamic strategy of episode resampling $E(\cdot)$ rather than fixed episodes of length $H$.
The proof presented in this paper attempts to apply the results of Lemma \ref{lem: posterior_sampling} for dynamically generated episode lengths as per Section \ref{sec: existing}.
This argument is quite appealing, but as we will show below, it is not rigorous and has made quite a delicate error.

The posterior sampling lemma (Lemma \ref{lem: posterior_sampling}) says that for any episode $k$, for any history $\Hc_k$ then the expected regret over one timestep:
\begin{equation}
\label{eq: epLen1}
   \Exp\left[ \lambda^*_* - \lambda^k_k \mid \Hc_{k1} \right] = 0.
\end{equation}
However, if the length of the episode $L^*_k$ depends on \textit{both} the true MDP $M^*$ and the sampled policy $\mu_k$ then in general,
\begin{equation}
\label{eq: epLen not one}
     \Exp\left[ L^*_k \left(\lambda^*_* - \lambda^k_k \right) \mid \Hc_{k1} \right] \neq 0.
\end{equation}
When we present the analyis using this notation, the statement of \eqref{eq: epLen not one} is quite clear and obvious.
However, this issue is quite subtle and can be quite hard to spot if we use a different notation for the problem.
In particular, this distinction between \eqref{eq: epLen1} and \eqref{eq: epLen not one} is at the root of an error in the proof of Theorem 2 in \cite{abbasi2015bayesian}.

We now reproduce the first steps of their proof (available at the bottom of page 15 in \url{https://arxiv.org/pdf/1406.3926.pdf}) using something more similar to their notation.
They write:
\begin{eqnarray*}
    \Exp \left[{\rm Regret}(T, \pi^{\rm LazyPSRL}, M^*) \right]
    &=& \sum_{t=1}^T \Exp\left[ \lambda^*_* - r_t \right] \\
    &=& \sum_{t=1}^T \Exp\left[ \Exp\left[ \lambda^*_* - r_t \mid \Hc_{k(t)1} \right] \right] \\
    (*) &=& \sum_{t=1}^T \Exp\left[ \Exp\left[ \lambda^{k(t)}_{k(t)} - r_t \mid \Hc_{k(t)1} \right] \right] \\
    &=& \sum_{t=1}^T \Exp\left[ \lambda^{k(t)}_{k(t)} - r_t \right],
\end{eqnarray*}
Where $k(t)$ to be the index of the active episode at time $t$ according to lazy PSRL.
In effect this argument is equivalent to $\sum_{t=1}^T \Exp\left[\lambda^*_* - \lambda^{k(t)}_{k(t)} \right] = 0$ for any $T$.
Unfortunately, and as we will now show this is not necessarily the case.

The authors in \cite{abbasi2015bayesian} justify step $(*)$ through an application of Lemma \ref{lem: posterior_sampling} since, conditional upon any $\Hc_{k(t)1}$, $\lambda^k_k$ is equal in distribution to $\lambda^*_*$.
Unfortunately, this argument for $(*)$ is not a valid application of Lemma \ref{lem: posterior_sampling}, since in general $k(t)$ is not $\sigma(\Hc_{k(t)})$-measurable.
This is because the process of episode signalling $E(\cdot)$ may depend on the history betwen $\Hc_{k(t)1}$ and $\Hc_t$, which is influenced by both $M^*$ and $\mu_k$.
We can relate this problem back to \eqref{eq: epLen not one} through an alternative decomposition,
\begin{equation}
    \sum_{t=1}^T \Exp\left[ \Exp\left[ \lambda^*_* \mid \Hc_{k(t)1} \right] \right]
    = \sum_{t=1}^{K(T)} \Exp\left[ \Exp\left[ L^*_k \lambda^*_* \mid \Hc_{k1} \right] \right]
    \neq \sum_{t=1}^{K(T)} \Exp\left[ \Exp\left[ L^*_k \lambda^k_k \mid \Hc_{k1} \right] \right].
\end{equation}
To clarify this point we will now consider a very simple explicit counterexample.

\begin{example}[Counterexample to step $(*)$]
\label{ex: counterexample}
\hspace{0.00001mm} \newline
Suppose the agent exists in a trivial MDP $M^*$ with $S=A=1$.
The rewards are deterministic and drawn from a prior with $R=1$ or $R=0$ with equal probability.
Consider a version of lazy PSRL with episode signal $E(\Hc_t) = \Ind \left\{ \sum_{j=1}^t r_t \ge 1 \ \text{\rm OR } \ t \ge H_{\rm max} \right\}$ for some $H_{\rm max} > 1$ and examine $\Exp[ \Delta^{\rm opt}_k ]$ for this agent.

Lemma \ref{lem: posterior_sampling} implies that, for any fixed dataset $\Hc_{k1}$, $\Exp[ \lambda^*_* | \Hc_{k1} ] = \Exp[ \lambda^k_k | \Hc_{k1}]$.
Suppose that the argument from $(*)$ were correct, this would imply that for all $T > 0$ and all $\Hc_{\rm max}$:
\begin{eqnarray*}
    \sum_{t=1}^T \Exp \left[ \lambda^*_* - \lambda^{k(t)}_{k(t)} \right]
    &=& \sum_{t=1}^T \Exp \left[\Exp \left[ \lambda^*_* - \lambda^{k(t)}_{k(t)} \mid \Hc_{k(t)1}  \right] \right]\\
    (*) &=& \sum_{t=1}^T \Exp \left[ \Exp \left[ \lambda^*_* - \lambda^*_* \mid \Hc_{k(t)1}  \right] \right] = 0
\end{eqnarray*}

However, we can clearly construct an example where this reasoning is false.
Fix $H_{\rm max} = 1000$ and consider $T = 1000$.
Then for this problem,
\begin{equation}
    \sum_{t=1}^{1000} \Exp \left[  \lambda^*_* -  \lambda^{k(t)}_{k(t)} \right]
    = \frac{1}{4} \times 1000 - \frac{1}{4} \times 1 > 249.
\end{equation}
This completes the counter-example to $(*)$ and the proof of Theorem 2 in \cite{abbasi2015bayesian}.
\end{example}

We note that the counterexample provided in Example \ref{ex: counterexample} demonstrates that the proof of Theorem 2 in \cite{abbasi2015bayesian} is incorrect.
However, the result of this Theorem \textit{may} still be true.
As such, we suggest that this result is considered as a conjecture, rather than established theorem.

%%%%%%%%%%%%%%%%%%%%%%%%%%%%%%%%%%%%%%%%%%%%%%%%%%%%%%%%%%%%%%%%%%%%%%%%%%%
%%%%%%%%%%%%%%%%%%%%%%%%%%%%%%%%%%%%%%%%%%%%%%%%%%%%%%%%%%%%%%%%%%%%%%%%%%%
\subsection{Pragmatic approaches to PSRL}
\label{sec: pragmatic}

Section \ref{sec: lazy_psrl} shows that there are some delicate issues in extending the analysis of PSRL to algorithms with dynamic episode resampling.
Despite these challenges, and the lack of a concrete theory in these settings, PSRL seems to do well even in settings without episodic reset \cite{Osband2013,abbasi2015bayesian}.
We end this technical note with a few pragmatic approaches to implement PSRL in these settings.

\begin{enumerate}
    \item For many problems we can impose a natural artificial episode length $H$, even if the underlying system is not strictly finite horizon.
    For example, the production levels in a power plant might be sensibly modeled as problem with $H= 1 {\rm day}$.
    Similarly, any problem which discounts future rewards by $\gamma \in (0,1)$ is quite similar to a finite horizon problem $H = O\left( \frac{1}{1-\gamma} \right)$.

    \item In the rare situation we are truly uncertain of the timeframe for the optimal policy $\mu^*$ then we can apply something similar to ``the doubling trick''.
    Although the results in Section \ref{sec: lazy_psrl} show that our current proof techniques are insufficient to maintain regret bounds in this setting, empirical evaluation seems to suggest this is not a problem for natural episode switching signals \cite{Osband2013,abbasi2015bayesian}.

    \item For some problems it may make more sense to \textit{slowly} vary the exploration noise gradually, rather than proceed in episodes of (potentially drastically) switching policies.
    In this setting we might separate the posterior samples for the MDP $M^*$ into some mean MDP $\hat{M}$ perturbed by some random noise $W$ such that $M^* = \tilde{f}(\hat{M}, W)$ for some $\tilde{f}_k$.
    In a system with discount rate $\gamma \in (0,1)$, we might consider an algorithm which resamples an MDP $M_t$ at \textit{every} timestep, but ensures that the MDP does not vary too quickly at each timestep.
    For example, if $P(s,a)$ is an unknown transition we might sample $P_t(s,a) \sim \phi(\cdot | \Hc_t)$ but then use a linearly smoothed estimate $\overline{P}_t := \gamma \overline{P}_{t-1}(s,a) + (1 - \gamma) P_t(s,a)$ to compute the policy $\mu_t$ which we actually follow.
\end{enumerate}

%%%%%%%%%%%%%%%%%%%%%%%%%%%%%%%%%%%%%%%%%%%%%%%%%%%%%%%%%%%%%%%%%%%%%%%%%%%
%%%%%%%%%%%%%%%%%%%%%%%%%%%%%%%%%%%%%%%%%%%%%%%%%%%%%%%%%%%%%%%%%%%%%%%%%%%
\section{Conclusion}

This technical note aims to clarify the current state of performance guarantees for reinforcement learning algorithms guided by posterior sampling.
In Section \ref{sec: lazy_psrl} we highlight the delicate technical issue which arises in the analysis of PSRL-variants in environments with dynamic episode length.
We highlight the mistake in the proof of Theorem 2 in \cite{abbasi2015bayesian} and reproduce a clear counterexample to show the mistake in this reasoning.
In Section \ref{sec: pragmatic} we suggest several pragmatic solutions to apply PSRL in environments without explicit episodic reset.
We believe that, in the future, it may be possible to extend existing analyses for finite horizon MDPs to a more general setting of learning without episodic reset.
We believe this is an interesting topic for future research and hope to stimulate further thinking on this topic.

%%%%%%%%%%%%%%%%%%%%%%%%%%%%%%%%%%%%%%%%%%%%%%%%%%%%%%%%%%%%%%%%%%%%%%%%%%%
%%%%%%%%%%%%%%%%%%%%%%%%%%%%%%%%%%%%%%%%%%%%%%%%%%%%%%%%%%%%%%%%%%%%%%%%%%%
\section*{Acknowledgements}

We would like to thank the authors of \cite{abbasi2015bayesian} for their help and dialogue in the discussion of these delicate technical issues.
We would also like to thank Daniel Russo for the many hours of discussion and analysis spent in the office on issues like these.
It is quite likely we would not have noticed the delicate technical issues of Section \ref{sec: lazy_psrl} were it not for his insights.

%%%%%%%%%%%%%%%%%%%%%%%%%%%%%%%%%%%%%%%%%%%%%%%%%%%%%%%%%%%%%%%%%%%%%%%%%%%
%%%%%%%%%%%%%%%%%%%%%%%%%%%%%%%%%%%%%%%%%%%%%%%%%%%%%%%%%%%%%%%%%%%%%%%%%%%
{
\small
\bibliographystyle{plainnat}
\bibliography{reference}
}

\end{document}